\documentclass[10pt,twocolumn,letterpaper]{article}

\usepackage{wacv}
\usepackage{times}
\usepackage{epsfig}
\usepackage{graphicx}
\usepackage{amsmath}
\usepackage{amssymb}
\usepackage{bm}
\usepackage{authblk}

\wacvfinalcopy 

\def\wacvPaperID{10} 

\ifwacvfinal\fi
\begin{document}

\title{An Order Preserving Bilinear Model for Person Detection in Multi-Modal Data}


\author[1]{Oytun Ulutan\thanks{ulutan@ece.ucsb.edu}}
\author[2]{Benjamin S. Riggan}
\author[3]{Nasser M. Nasrabadi}
\author[1]{B. S. Manjunath}
\affil[1]{Department of Electrical and Computer Engineering, University of California, Santa Barbara, CA}
\affil[2]{US Army Research Lab, Adelphi, MD}
\affil[3]{West Virginia University, Morgantown, WV}

\renewcommand\Authands{ and }

\maketitle
\ifwacvfinal\thispagestyle{empty}\fi

\begin{abstract}
We propose a new order preserving bilinear framework that exploits low-resolution video for person detection in a multi-modal setting using deep neural networks. In this setting cameras are strategically placed such that less robust sensors, e.g.~geophones that monitor seismic activity, are located within the field of views (FOVs) of cameras. The primary challenge is being able to leverage sufficient information from videos where there are less than 40 pixels on targets, while also taking advantage of less discriminative information from other modalities, e.g. seismic. Unlike state-of-the-art methods, our bilinear framework retains spatio-temporal order when computing the vector outer  products between pairs of features.  Despite the high dimensionality of these outer products, we demonstrate that our order preserving bilinear framework yields better performance than recent “orderless” bilinear models and alternative fusion methods. Code is available at https://github.com/oulutan/OP-Bilinear-Model
\end{abstract}

\section{Introduction}


Human detection is a frequently studied problem, especially in the context of surveillance applications \cite{bahrampour2013performance, damarla2007personnel, damarla2011detection, sabatier2008range}.  In our work, we are interested in cases where visual detectors fail due to insufficient number of pixels on the target (i.e., low resolution).  Therefore, our objective is to provide a detection framework that is robust to challenging conditions, such as few pixels on target, by leveraging multi-modal sensor data.


Low-resolution videos can be generated from a scenario where a high resolution camera with a wide field of view (FOV) placed close to a power source but far away from the field with targets. This requires visual detection frameworks to search for small (few pixels) objects on a large field. Seismic sensors on the other hand can provide reliable information about their close surroundings and can easily be distributed on a large field. This allows the data from a seismic sensor to improve the detection of cameras in regions where camera view and sensor range intersects.

\begin{figure}[t]
\begin{center}
\includegraphics[width=1.0\linewidth]{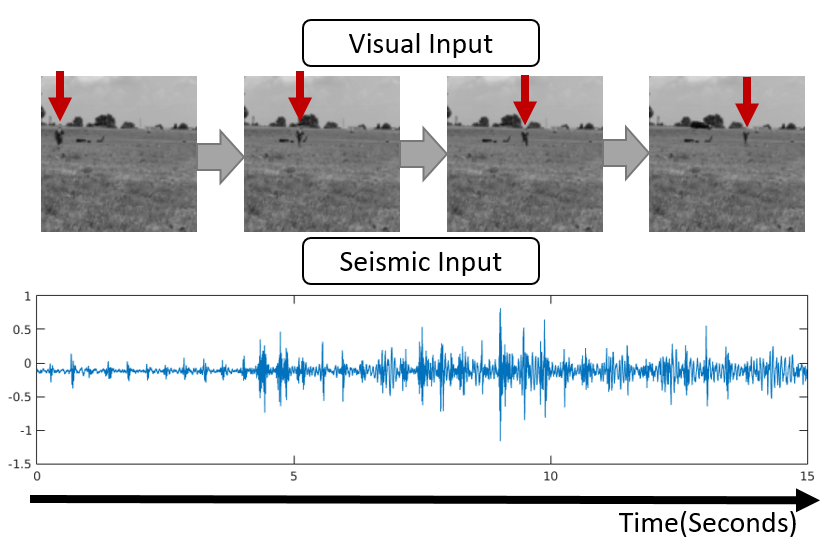}
\end{center}
   \caption{An example of time synchronized seismic and visual data. Frames are cropped centering the seismic sensor's location. As a person gets closer to the center of image, the amplitude of the seismic signals increase. Red arrows indicate the person.}
\label{fig:problem}
\end{figure}

In this work, we consider a typical surveillance setting (e.g., border patrol) where multiple sensors and cameras are used to monitor a particular area.  Traditional methods for person detection that rely only upon visual cues tend to perform poorly on low resolution imagery data from our dataset. For this reason, we aim to jointly leverage corresponding sensor (e.g., seismic) and imaging data (Fig.~\ref{fig:problem}).

In this context, we propose a new order-preserving bilinear fusion model for person detection, leveraging pairwise interactions between convolutional features in a new way. We demonstrate that sparse feature selection combined with bilinear fusion selects the optimal combinations of spatio-temporal features. We show that the proposed fusion method is differentiable and the final model is end-to-end trainable. The performance of our fusion model is tested in a new multi-modal person detection dataset with syncronized seismic sensors and video cameras \cite{nabritt2015personnel}. The dataset is available through requests\footnote{The dataset can be obtained by sending an email to \newline benjamin.s.riggan.civ@mail.mil }. 
Our experimental results show that our model achieves better detection accuracy and reduced false positive rates compared to the state of the art fusion methods. 

\section{Related Work}

In a surveillance setting, traditional detection methods for multimodal sensor data depend on hand-crafted features such as frequency domain analysis \cite{damarla2011detection, sabatier2008range}, Symbolic Dynamic Filtering \cite{bahrampour2013performance}, and Cepstral features \cite{nguyen2011robust}. Damarla \etal\cite{damarla2007personnel} extracts and fuses hand-crafted features from multiple different modalities for person detection. Recently, with the advances of computational hardware and the increase of available data, feature learning has been integrated with classification to achieve end-to-end trainable systems \cite{krizhevsky2012imagenet}.

Ngiam \etal \cite{ngiam2011multimodal}  analyzed the relations between different modalities in deep networks and showed that cross-modality feature learning can improve single modality performance. Riggan \etal \cite{riggan2015coupled} used Coupled AutoEncoders for cross-modal face recognition fusing visible and thermal imaging. \cite{eitel2015multimodal, socher2012convolutional} achieved fusion by concatenating features from CNNs trained on RGB and depth images.

Fusing different features extracted from a single modality has been achieved using multiple different methods which are also applicable to multi-modal fusion. \cite{simonyan2014two, wang2015towards} achieved late fusion between optical flow and RGB by averaging the confidence scores of single CNNs for video classification. Karpathy \etal \cite{karpathy2014large} analyzed concatenating features from different time instances and trained fully connected layers to fuse information over time in a video. 

Bilinear models were first analyzed by Tenenbaum and Freeman \cite{tenenbaum2000separating} to manipulate two factors from images, style and content. Recently bilinear models have achieved success in multiple tasks. Lin \etal \cite{lin2015bilinear} fused two convolutional neural networks to obtain orderless descriptors and improved results in fine-grained visual recognition. Carreira \etal \cite{carreira2012semantic} used second order statistics of the local descriptors for semantic segmentation. RoyChowdhury \etal \cite{roychowdhury2015face} used bilinear CNNs to improve results in face identification tasks. Gao \etal \cite{Gao_2016_CVPR} improves the bilinear methods by developing a compact pooling method.

The main difference between recent bilinear methods \cite{carreira2012semantic, Gao_2016_CVPR, lin2015bilinear, roychowdhury2015face} and our method is that we use the outer product of vectors and obtain the pairwise feature interactions at each spatio-temporal indices. This is in contrast with these methods that use pooling methods over all indices and obtain an `orderless' descriptor without preserving the order. 




\section{Technical Approach}


\begin{figure}[t]
\begin{center}
\includegraphics[width=0.9\linewidth]{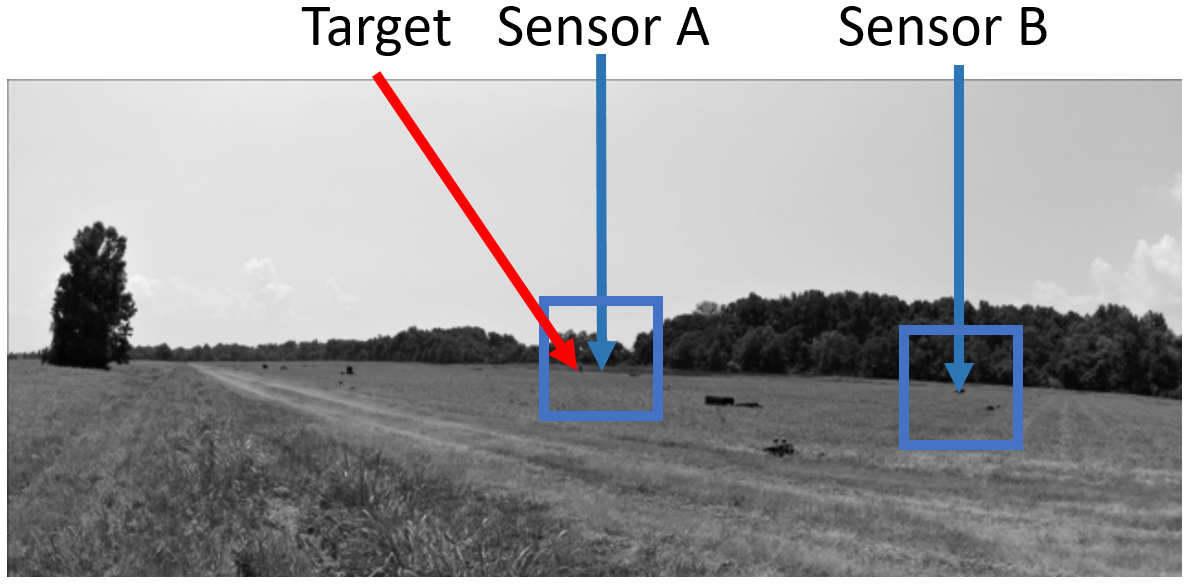}
\end{center}
   \caption{ROIs seen from a wider camera view. Each ROI is located around the sensor locations which are known a priori. Notice that in the figure, target is within the Sensor A's region which produces a positive sample whereas the sample from Sensor B is a negative sample. There are multiple ROIs within this camera frame but only two of them are shown.}
\label{fig:bbox}
\end{figure}


The goal is to detect the region of interest(ROI) with a person walking in a field that is being monitored by a multi-modal sensor network data consisting of video cameras and seismic geophones. In this context, a ROI is any contiguous set of pixels and corresponding sensor data. Detection is defined on ROIs with corresponding camera and sensor pairs. We pose this as a binary classification problem for each ROI. Fig.~\ref{fig:bbox} shows example ROIs located around the sensor locations which are known a priori. The inputs to our model are a single optical flow frame and its corresponding seismic signal for the same time interval. 

In the following sections, we define the problem as a general multi-modal fusion problem and derive our fusion model by explaining each of the modules. 

\subsection{Problem Definition} \label{sec:problem}

\begin{figure*}[t]
\begin{center}
\includegraphics[width=0.9\linewidth]{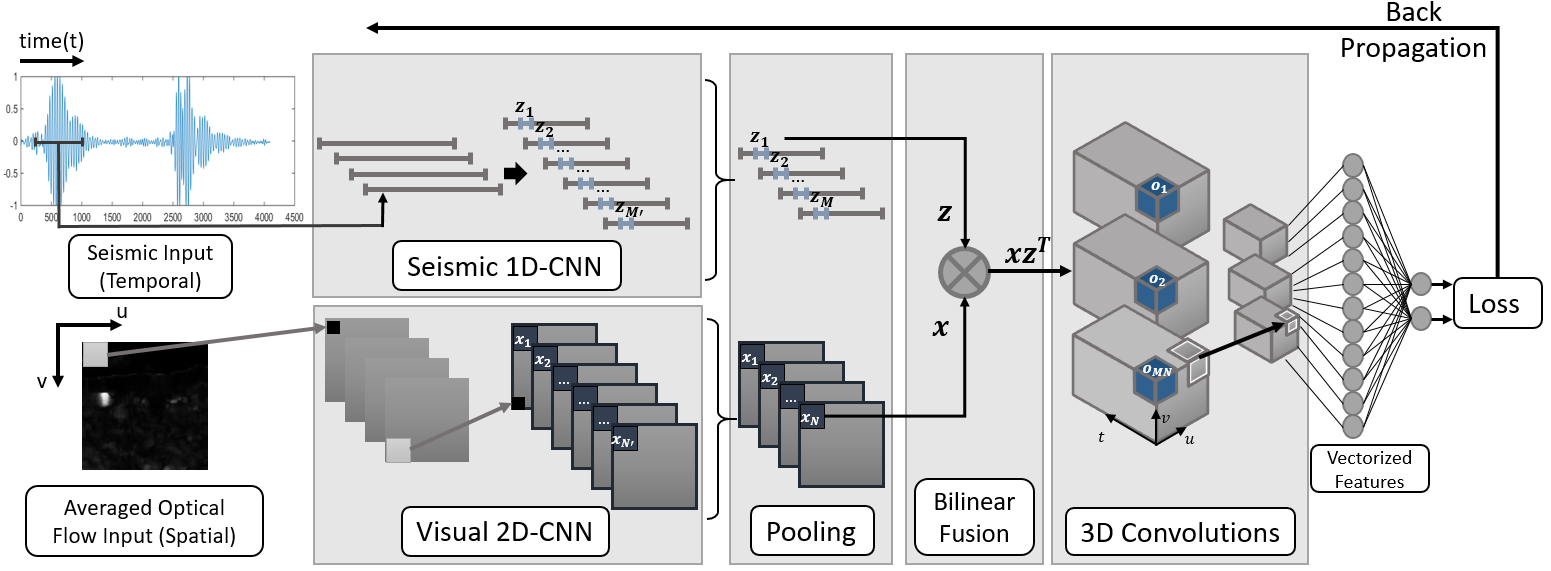}
\end{center}
   \caption{Order Preserving Bilinear model: Data from both modalities go through their respective CNN streams. Resulting features are compressed into lower dimensional vectors by sparse feature reduction and then fused by taking outer product at every spatio-temporal index. Since the order is preserved, 3D convolutions are leveraged. Since every module is differentiable, the whole model is trained end-to-end. }
\label{fig:model}
\end{figure*} 

Let  $\mathit{X}$ and $\mathit{Z}$ be two sets of local descriptors extracted from two different modalities. Each descriptor $\bm{x}_{u_x, v_x, t_x} \in \mathit{X}$ represents the feature vector for the spatio-temporal voxel defined by the indices $u_x, v_x, t_x$, and similarly for the other modality $\bm{z}_{u_z, v_z, t_z} \in \mathit{Z}$. Let $\bm{x}$ and $\bm{z}$ be $N\times1$ and $M\times1$ dimensional feature vectors respectively. 

Our goal is to develop a fusion algorithm  $\mathit{O} = f(\mathit{X},\mathit{Z})$ such that spatio-temporal indices are preserved. For every spatio-temporal index from both modalities, we have the output feature vector:

\begin{equation}
    \bm{o}_{u_x, v_x, t_x, u_z, v_z, t_z} = f(\bm{x}_{u_x, v_x, t_x}, \bm{z}_{u_z, v_z, t_z})
\label{eq:output_long}
\end{equation}

\noindent where $\bm{o}_{u_x, v_x, t_x, u_z, v_z, t_z} \in \mathit{O}$ are the local descriptors of the output. If the input modalities are synchronized in time and space then we will have $(u_x, v_x, t_x) = (u_z, v_z, t_z) = (u,v,t)$. Indices from Eq. \ref{eq:output_long} simplifies into:

\begin{equation}
    \bm{o}_{u, v, t} = f(\bm{x}_{u, v, t}, \bm{z}_{u, v, t})
\label{eq:output_short}
\end{equation}

Furthermore, if we let modality $\mathit{X}$ to be a spatial signal and modality $\mathit{Z}$ to be a temporal signal. That gives $t_x = 1$ and $u_z = v_z = 1$ and simplifies the Eq. \ref{eq:output_long} into:

\begin{equation}
    \bm{o}_{u, v, t} = f(\bm{x}_{u, v}, \bm{z}_{t})
\label{eq:output_st}
\end{equation}

Eq. \ref{eq:output_st} defines the local descriptor which is the output of the fusion method. Note that in both Eq. \ref{eq:output_short} and Eq. \ref{eq:output_st}, the calculation of $\bm{o}_{u, v, t}$, depends on the input values at indices $u, v, t$, which gives an ordered descriptor. Ordered descriptors allow us to exploit the relations between neighboring terms by using methods such as 3D convolutions. The goal is to detect targets using spatial images and temporal seismic sensor data, which fits into the formulation in Eq.~\ref{eq:output_st}.

Our model is organized into four sub-components as shown in Fig.~\ref{fig:model}: \textbf{1)} input sensor signals are processed by dedicated CNNs for each modality (Section \ref{sec:CNNs}); \textbf{2)} at each spatial and temporal index, feature vectors are compressed in their depth dimension (Section \ref{sec:pooling}); \textbf{3)} outer product is used in each spatio-temporal index to obtain the bilinear feature vector (Section \ref{sec:bilinear_fusion}); and \textbf{4)} 3D convolutions are used to leverage neighborhood relations of spatio-temporally ordered terms (Section \ref{sec:3dconv}).


\subsection{CNN Features} \label{sec:CNNs}

In previous works, CNNs have been shown to extract useful features for variety of tasks on spatial \cite{He_2016_CVPR}, temporal \cite{abdel2014convolutional} and spatio-temporal \cite{simonyan2014two} modalities. CNNs extract local feature vectors at each spatio-temporal index $(u,v,t)$. The size of the vector depends on the number of filters in the last convolutional layer, i.e., depth of the layer. For each modality at each index $u,v,t$ we have:

\begin{equation}
    \bm{x'}_{u, v} = [ x'_1, x'_2, ... x'_{N'}]^T,
\label{eq:feats_x}
\end{equation}

\begin{equation}
    \bm{z'}_{t} = [ z'_1, z'_2, ... z'_{M'}]^T.
\label{eq:feats_z}
\end{equation}

\noindent The prime ($'$) notations refer to the values before feature selection.

\subsection{Sparse Feature Selection}\label{sec:pooling}

The proposed fusion method, explained in Section \ref{sec:bilinear_fusion}, generates a high dimensional vector. Using high dimensional vectors are computationally challenging and can be prone to overfitting due to increased number of parameters. Within these large number of features, we want to prioritize which feature pairs are more useful (further discussed in Section \ref{sec:effects}). Therefore, we implement an efficient way to perform spatio-temporal feature selection by combining sparse $1\times1$ convolutions with bilinear fusion.  Moreover, this method maintains spatio-temporal order. The goal is to compress the input vector to reduce the dimensions from Eq. \ref{eq:feats_x}. From here on, we generically use the term `\textbf{reduction}' to represent both feature selection and dimensionality reduction operations. We define our reduction function $r(.)$ as:

\begin{equation}
    r(\bm{x'}_{u, v}) = \bm{x}_{u, v} = [ x_1, x_2, ... , x_{N}]^T,
\label{eq:pool_x_vector}
\end{equation}

\noindent where $N<N'$ so that we obtain a more compact feature vector and we define the each reduced component $x_i$ as the linear combinations of the original vector:

\begin{equation}
    x_i = ReLU(\sum_{k=1}^{N'}w^x_{ik}  x'_k) = max(0,\sum_{k=1}^{N'}w^x_{ik}  x'_k),
\label{eq:pool_x}
\end{equation}

\noindent where weights $w^x_{ik}$ are learned over the training and the norm of the weights are regularized using $L1$ normalization. Compared to $L2$ normalization or without normalization, $L1$ normalization generates a more sparse set of weights which forces the network to `choose' the features that will be included in the summation. By $L1$ regularization, the weights $|w^x_{ik}|$ are mostly close to zero except a few weights that are multiplying essential set of features $x'_k$. This is similar to LASSO \cite{yuan2006model,meier2008group} and provides a feature selection operation. Similarly for the second modality, reducing the vector from Eq. \ref{eq:feats_z}:

\begin{equation}
    r(\bm{z'}_{t}) = \bm{z}_{t} = [ z_1, z_2, ... z_{M}]^T,
\label{eq:pool_z_vector}
\end{equation}


\begin{equation}
    z_i = ReLU(\sum_{k=1}^{M'}w^z_{ik}  z'_k) = max(0, \sum_{k=1}^{M'}w^z_{ik}  z'_k).
\label{eq:pool_z}
\end{equation}

\subsection{Order Preserving Bilinear Fusion}\label{sec:bilinear_fusion}

Reduced CNN features (Eq. \ref{eq:pool_x_vector} and Eq. \ref{eq:pool_z_vector}) are fed into the fusion layer. At each spatial and temporal index, local feature vectors from both modalities are fused by taking the outer product. The fusion function at each spatio-temporal index $u \in U, v \in V, t \in T$ can be written as: 
\begin{equation}
    \bm{o}_{u, v, t} = f(\bm{x}_{u, v}, \bm{z}_{t}) = vectorize(\bm{x}_{u, v}\bm{z}_{t}^T)
\label{eq:fused_feats}
\end{equation}

At each index, we have length $N$ vector $\bm{x_{u,v}}$ and length $M$ vector $\bm{z_t}$. Outer product between these feature vectors generate the $N \times M$ second order pairwise features matrix:

\begin{equation}
    \bm{x}_{u, v}\bm{z}_{t}^T =
    \begin{bmatrix}
        x_1z_1 & x_1z_2 & ...    & x_1z_M \\
        x_2z_1 & x_2z_2 & ...    & x_2z_M \\
        \vdots   &          & \ddots & \vdots \\
        x_Nz_1 & x_Nz_2 & ...    & x_Nz_M \\
    \end{bmatrix}.
\label{eq:pair_mtx}
\end{equation}

We stack the rows together in lexicographical order, i.e., $N\times M$ dimensional matrix into an $MN\times1$ vector. This gives the fused feature vector at each spatio-temporal index. 

\begin{equation}
\begin{aligned}
    &\bm{o}_{u, v, t} = [o_1, o_2, ... o_{MN}]^T = \\
    &\begin{bmatrix}
        x_1z_1  &... & x_1z_M & ...  & x_Nz_1  & ...    & x_Nz_M
    \end{bmatrix}^T
\end{aligned}
\label{eq:fused_vector}
\end{equation}

We repeat this operation for each spatial index $u,v$ and temporal index $t$ and obtain the fused second order feature vector at every combination of indices $u,v,t$.

\subsubsection{Differentiability for Backpropagation}\label{sec:diff}
This fusion operation is differentiable for gradient operations and it is end-to-end trainable. In this section we show how the gradient can be backpropagated to each modality stream. Let $L$ denote the cross-entropy loss function. Then by chain rule, we obtain:

\begin{equation}
\scriptsize
    \frac{\partial L}{\partial \bm{x}_{u, v}} = \frac{\partial L}{\partial \bm{o}_{u, v,t}}\frac{\partial \bm{o}_{u, v,t}}{\partial \bm{x}_{u, v}} =  \frac{\partial L}{\partial \bm{o}_{u, v,t}}
    \begin{bmatrix}
        \frac{\partial o_1}{\partial x_1}  & ...    & \frac{\partial o_1}{\partial x_N} \\
        \vdots  &  \ddots  & \vdots \\
        \frac{\partial o_{MN}}{\partial x_1}  & ...    & \frac{\partial o_{MN}}{\partial x_N} \\
    \end{bmatrix}
\label{eq:gradientx}
\end{equation}

\noindent where $\frac{\partial L}{\partial \bm{o}_{u, v,t}}$ can be calculated using chain rule of derivatives for layers between loss $L$ and the outer product. Each partial derivative in the matrix can be written as:

\begin{equation}
    \frac{\partial o_p}{\partial x_r} = \frac{\partial (x_sz_q)}{\partial x_r}
\label{eq:gradient_op}
\end{equation}

\noindent where $p=1,..,MN$, $q=1,..,M$, $r=1,..,N$ and $s=1,..,N$. For $s=r$, this simplifies into:

\begin{equation}
    \frac{\partial o_p}{\partial x_r} = \frac{\partial (x_rz_q)}{\partial x_r} = z_q
\label{eq:gradient_zop}
\end{equation}

\noindent For $s \neq r$, Eq. \ref{eq:gradient_op} becomes 0. Gradients before this layer can also be calculated by the regular CNN chain rule. $\frac{\partial L}{\partial\bm{z}_{t}}$ can be calculated similarly for the second modality.

\subsubsection{Effects of Feature Selection}\label{sec:effects}

The outer product generates a high dimensional feature vector at each index $\bm{o}_{u, v, t}$. To handle the high dimensionality, we pool the convolutional features before the outer product operation by feature selection (Section \ref{sec:pooling}).
When $\sum_{k=1}^{N'}w^x_{ik}  x'_k > 0$ and $\sum_{l=1}^{M'}w^z_{jl}  z'_l > 0$, multiplying the terms from Eq.~\ref{eq:pool_x} and Eq.~\ref{eq:pool_z} yields for each $x_i z_j$ in Eq.~ \ref{eq:fused_vector}:

\begin{equation}
    x_i z_j = \sum_{k=1}^{N'}w^x_{ik}  x'_k \times \sum_{l=1}^{M'}w^z_{jl}  z'_l = \sum_{k=1}^{N'}\sum_{l=1}^{M'}w^{xz}_{kl} x'_k z'_l
\label{eq:pool_fuse}
\end{equation}

\noindent where each $w^{xz}_{kl} = w^x_{ik}w^z_{jl}$. Otherwise, the $x_i z_j = 0$. This shows that output of reduced fusion operation is linear combinations of the second order interactions of the original feature vectors before the feature selection operation $x'_k, z'_l$.

Weights of $1 \times 1$ convolutions $w^x_{ik}, w^z_{jl}$ are trained with $L1$ regularization, hence they are individually sparse (Section \ref{sec:pooling}). Therefore, this ensures that when multiplied, the produced set of weights are also sparse and the product $w^x_{ik} w^z_{jl}$ is non-zero only if corresponding features $k,l$ from each modality $x'_k, z'_l$ are individually important for the task which is similar to sparse representations\cite{patel2011sparse}.

\subsection{3D Convolutions}\label{sec:3dconv}

Since the outer product operation is repeated for every combination of the spatial ($u,v$) and temporal ($t$) indices, output of the fusion operation is a spatio-temporal feature tensor as shown in Fig.~\ref{fig:model}. This tensor allows us to use shared weights that stride across spatial and temporal dimensions, i.e., 3D convolutions, to reduce the total number of parameters and chances of overfitting by exploiting spatio-temporal correlations. In the tensor, at every spatio-temporal index, we have a feature vector of length $MN$ which is the output of the outer product between length $M$ vector $\bm{x}$ and length $N$ vector $\bm{z}$. In the 3D convolutions, this dimension corresponds to the depth of input. The intuition behind keeping the spatio-temporal order is that certain activations in certain combinations of spatial and temporal indices complement each other. By having all the second order pairs as features at each index, we can find feature pairs that are sufficiently discriminative.

\section{Implementation Details}

The data is collected in a sensor field with 16 seismic sensors and 4 video cameras\cite{nabritt2015personnel}. Seismic sensors are placed on a grid and the video cameras are placed outside the sensor field, observing it from different directions. 
In a surveillance setting, viewpoints and conditions vary for cameras and sensors, and surroundings can change the detected signature of the seismic sensors. To take this into account and to make the model generalizable, we split the data such that camera views (angle, background) and seismic sensors that are used in test set are different than the ones in training set. Each person in the field wears a GPS sensor. Using the location information we label the samples as positive when a person is within 15 meters of a seismic sensor. This results in 69483 negative and 16481 positive samples in training set and 26064 negative and 6440 positive samples in test set.

Videos are recorded at 30 frames per second at $640\times360$ resolution and seismic signals captured at 4096 Hz sampling rate. A $100\times100$ region that is centered at a seismic sensor location (known a priori) is cropped from each camera frame. From seismic signals we extract our data points as 1 second intervals with 50\% overlap. 
For the video data, we compute optical flow(OF). For each seismic signal centered at time $t$, OF frames are computed from the seismic sensor's corresponding region over the time interval $[t-1,t+1]$. Magnitudes of these OF frames are averaged and used as the input to the proposed method. By averaging OF frames the spatio-temporal modality video is compressed into a spatial representation that encodes the temporal motion information. The reasoning behind this approach is mostly computational. This approach is further investigated and compared to LSTMs in Section \ref{sec:lstm}.

To measure the performance of our methods, we report the precision, recall and F1-score values for the positive class. Recall values measure the detection accuracy whereas Precision measures the rate of false positives. In a data as unbalanced as ours, reporting both recall and precision becomes important. Since the negative class has significantly more samples than the positive class, high accuracy in detecting negative samples might still mean high false positive rates. For example $90\%$ accuracy in negative test samples still means $26064 \times 0.10 = 2606$ false positives which is $40\%$ of the total number of positive samples. 

All models are trained using TensorFlow \cite{tensorflow2015-whitepaper} and optimized using ADAM optimizer\cite{kingma2014adam}.

\subsection{Single Modality CNNs}\label{sec:single_CNN}

For extracting useful features from both seismic and visual data, we independently train modality specific CNNs for the detection task and analyze their performances. 

Since there are no similar works using seismic sensors to be used for transfer learning, a randomly initialized 1-dimensional CNN is trained for the seismic modality. For the visual modality, we leverage the Inception V3 network architecture explained in \cite{szegedy2015rethinking} and initialize the network with weights that are pretrained for ImageNet \cite{russakovsky2015imagenet}. Since this network is trained on RGB images and trained to detect ImageNet-specific features, we use earlier layers instead of the full architecture. Earlier layers in a CNN extract basic features such as edges, corners and these features are more generalizable. In \cite{NIPS2014_5347} the authors quantified the generality and specificity of the layers and showed that the earlier layers are more generalizable. In \cite{WangXWQLTV16} an OF CNN for action recognition is initialized using weights from a model trained for ImageNet. In our case, for the OF CNN we use the first five convolutional layers from Inception V3 model and initialize the weights from a ImageNet trained model.

\subsection{Order Preserving Bilinear Fusion}

The proposed approach (Fig.~\ref{fig:model}) consists of two dedicated streams of CNNs for each modality (Section \ref{sec:CNNs}), their corresponding sparse feature selection layers (Section \ref{sec:pooling}), outer product between outputs of the two streams at each spatio-temporal index to preserve the order (Section \ref{sec:bilinear}), 3D convolutions (Section \ref{sec:3dconv}) and a final fully connected layer for classification. We refer this model as Order Preserving (OP) Bilinear Model.

Architectures used for the modality dedicated CNN streams are the same architectures as the single modality models defined in previous section. This allows us to initialize the model weights with pretrained weights from single modality models. Each CNN stream is followed by sparse feature selection and the fusion is achieved by order preserving outer product operation. Since the proposed outer product fusion is differentiable, as shown in Section \ref{sec:diff}, the whole model is  fine-tuned in an end-to-end fashion.

\section{Experiments and Results}



In the following sections, we conduct a series of experiments to analyze the performance of each module in our method. First, we report experiments on the single modality CNNs and analyze the effects of dimensionality reduction. Then, we demonstrate the superior performance of the proposed bilinear fusion method compared to single modality models and alternative fusion methods. Furthermore, we compare the order-preserving methods that exploit 3D convolutions with their fully connected counterparts. Finally, we compare our visual approach with a LSTM approach.




\subsection{Impact of Sparse Feature Reduction}


For each modality, two different models are trained. Initial models use convolutional layers followed by fully connected layers. These models are labeled as `Seismic' and `Visual' in the tables. Additionally, we train models with the sparse feature selection method explained in Section \ref{sec:pooling}. We add the feature selection layer between convolutional and fully connected layers. These models are labeled as `Seismic Reduced' and `Visual Reduced' in the tables.

Table \ref{tab:acc_results} implies that sparse feature selection (reduced models) from Section \ref{sec:pooling} provide a slight trade-off in performance for computation efficiency for computing bilinear features. In the Visual CNN, the reduction in number of parameters are significant with this reduction method.

\begin{table}
\begin{small}
\begin{center}
  \begin{tabular}{| l || c | c | c | }
    \hline
    Model            & Recall & Precision & F1-Score\\ \hline \hline
    Seismic          & 0.90 & 0.87 & 0.89 \\ \hline
    Seismic Reduced  & 0.89 & 0.86 & 0.88 \\ \hline
    Visual          & 0.82 & 0.89 & 0.86 \\ \hline
    Visual Reduced   & 0.78 & 0.89 & 0.83 \\ \hline
    \textbf{OP-Bilinear Fusion} & \textbf{0.97} & \textbf{0.96} & \textbf{0.96} \\ \hline
    
  \end{tabular}
\end{center}
\caption{Precision, Recall and F1-Score values for single modality models and the proposed fusion method.}
\label{tab:acc_results}
\end{small}
\end{table}

\subsection{Fusion Compared to Single Modalities}\label{sec:bilinear}

\begin{table}
\begin{footnotesize}
\begin{center}
  \begin{tabular}{| l || c | c | c | }
    \hline
    Distances From \textbf{Cameras}(meters) & 50-80  & 80-110 & 110-140 \\ \hline \hline
    Visual Reduced & 0.96 & 0.93 & 0.74 \\ \hline
     \textbf{OP-Bilinear Fusion}    & \textbf{0.98} & \textbf{0.96} & \textbf{0.95} \\ \hline\hline \hline
    Distances From \textbf{Sensors}(meters)    & 0-5  & 5-10 & 10-15 \\ \hline \hline
    Seismic Reduced & 0.96 & 0.93 & 0.80 \\ \hline
     \textbf{OP-Bilinear Fusion}    & \textbf{0.99} & \textbf{0.97} & \textbf{0.93} \\ \hline
    
  \end{tabular}
\end{center}
\caption{Recall rates for different distances from the cameras and seismic sensors. Even though the performance of OP-Bilinear model also decreases with range, the change is not as significant since it incorporates the information from the complementary modality.}
\label{tab:bil_vs_distance}
\end{footnotesize}
\end{table}

Table \ref{tab:acc_results} compares the proposed fusion method against single modality models and shows that the fusion method provides the best performance in accuracy (Recall) and false positive rate (Precision). 
Fig.~\ref{fig:prec_recall_deep} compares the method with other select models by plotting Precision-Recall curves. This plot demonstrates that our model is the best performing classifier since OP-Bilinear curve achieves the best Precision-Recall trade-off at every point. 

\begin{figure}[t]
\begin{center}
\includegraphics[width=1.0\linewidth]{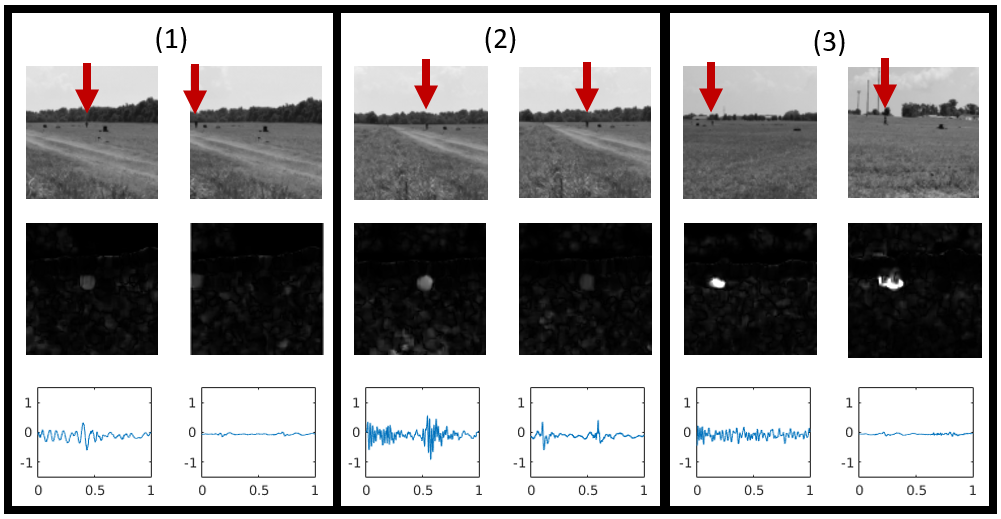}
\end{center}
   \caption{Examples of correct detections from the OP-Bilinear Model where single modality models fail. Red arrows indicate the targets.}
\label{fig:fusion_gained}
\end{figure}

Fig.~\ref{fig:fusion_gained} shows 3 sets of data samples. The first set shows the cases where both Visual and Seismic models fail but the fusion model correctly detects the target. In both samples, OF captures a weak motion and seismic sensor captures noise-like signals, but the fusion method detects the person nevertheless. The second set shows the samples where Visual model fail but Seismic and OP-Bilinear models correctly detects the target. Similarly, the third set shows the samples where Seismic model fails but Visual and OP-Bilinear model detects the target. This demonstrates that the fusion model achieves robust detection even when the input from a single sensor deteriorates.

We further compare the fusion model to the single modality models. As the distance between the target and the sensors increase, the performance deteriorates. Table \ref{tab:bil_vs_distance} demonstrates that the proposed OP-Bilinear Fusion model is more robust to distance. The fusion model can effectively incorporate the information from the complementary modality when one modality degrades with range.

\subsection{Effects of Initialization}

\begin{table}
\begin{small}
\begin{center}
  \begin{tabular}{| l || c | c | c | }
    \hline
    Model            & Recall & Precision & F1-Score\\ \hline \hline
    End-to-End OP-Bilinear & 0.95 & 0.95 & 0.95 \\ \hline
    \textbf{OP-Bilinear Fusion} & \textbf{0.97} & \textbf{0.96} & \textbf{0.96} \\ \hline
    
  \end{tabular}
\end{center}
\caption{Precision, Recall and F1-Score values for different initialization methods. }
\label{tab:bilinear_variants_results}
\end{small}
\end{table}

In Section \ref{sec:diff}, we have derived the gradient for the proposed outer product operation. Since the gradient exists, the whole model is end-to-end trainable. In the previous section, we showed the results of the proposed method by initializing the model with single modality CNN model weights and fine-tuning the whole model. To investigate end-to-end training, we train a model using the same architecture, except the filter weights for the model are randomly initialized. Table \ref{tab:bilinear_variants_results} compares the performance of the end-to-end trained network with the model that is fine-tuned on pre-trained weights. This shows that pre-training achieves a slightly better performance than random initialization.

    

\subsection{Comparisons with Fusion Methods}

\begin{figure}[t]
\begin{center}
\includegraphics[width=0.9\linewidth]{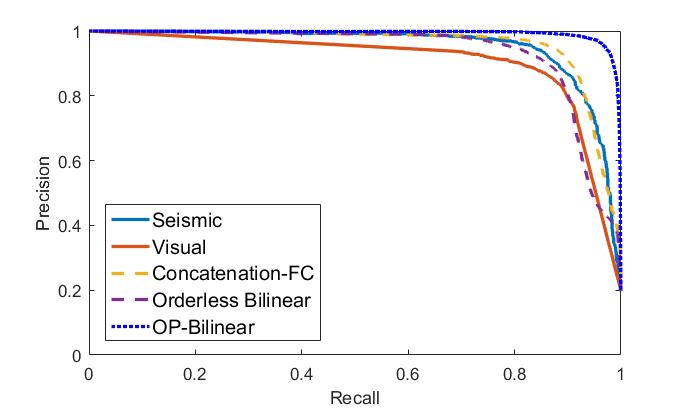}
\end{center}
   \caption{Precision-Recall curves show that OP-Bilinear Fusion achieves the best detection rate and fewest false positives.} 
\label{fig:prec_recall_deep}
\end{figure}

We compare our proposed OP-Bilinear Model with multiple late fusion approaches, feature concatenation approaches and state of the art Orderless Bilinear methods. 


\textbf{Average Fusion: }We compare our results with a simple confidence score averaging late fusion method. This is a widely used method due to its simplicity \cite{karpathy2014large, simonyan2014two, wang2015towards}. In this method, we take the confidence scores from individually trained models `Seismic' and `Visual' from Section \ref{sec:single_CNN} and average them to get the final score for each datapoint. Results are labeled as `Average Fusion' in Table \ref{tab:fusion_results}.


\textbf{Dempster Shafer Fusion: }We compare our results with a more sophisticated late fusion method, Demster Shafer theory \cite{dempster1967upper}. This theory is a framework for reasoning with uncertainty and generally applicable to sensor fusion models. We implement this model similar to \cite{lee2017fusion}. We assume more uncertainty for visual modality than seismic modality, e.g. 35\% versus 15\%, due to noise and resolution. The results of this framework are shown in Table \ref{tab:fusion_results} with label `Demster Shafer Fusion'.

Compared with these late fusion methods, our proposed fusion method is able to model the relations between the modalities and achieve better performance. Table \ref{tab:fusion_results} demonstrates that the proposed OP-Bilinear fusion model achieves higher detection rate (Recall) with lower false positive rate (Precision). 


\textbf{Concatenation-Fully Connected: }Many multi-modal fusion \cite{eitel2015multimodal, socher2012convolutional, wagner2016multispectral} and feature fusion \cite{karpathy2014large} methods concatenate the feature vectors from CNNs and classify the results using fully connected layer. This simple stacking of feature vectors compresses the spatial or temporal order since the features at every index are stacked into a single vector. Note that such operation does not exploit correlations in the spatial or temporal order. 
The output of this fusion can be expressed as:

\begin{equation}
\begin{aligned}
    &\bm{o}_{u, v, t} = [o_1, o_2, ... o_{M+N}]^T = \\
    &\begin{bmatrix}
        x_1  &... & x_N & z_1 &  ...    & z_M
    \end{bmatrix}^T
\end{aligned}
\label{eq:concat_vector}
\end{equation}

\noindent and vectors at each spatial and temporal indices are also stacked into a vector as:

\begin{equation}
\begin{bmatrix}
        \bm{o}_{1, 1, 1}  &... & \bm{o}_{u, v, t} &  ...    & \bm{o}_{U, V, T}
    \end{bmatrix} ^T
\label{eq:concat_os}
\end{equation}

Results of this model are provided in Table \ref{tab:fusion_results} and Fig.~\ref{fig:prec_recall_deep} under the label `Concatenation-FC'. The results show that the OP-Bilinear method achieves better performance than the Concatenation model by extracting bilinear features and preserving order. 


\textbf{Orderless Bilinear Descriptor: }Bilinear pooling methods \cite{lin2015bilinear, roychowdhury2015face, carreira2012semantic, Gao_2016_CVPR} use sum pooling over spatial indices to pool the second order feature tensor into an orderless feature representation. Inspired by this idea, we sum the output of the outer product operation $\bm{x}_{u, v}\bm{z}_{t}^T$ from every spatial and temporal indices.

\begin{equation}
    \sum_{u,v,t}\bm{x}_{u, v}\bm{z}_{t}^T =
    \begin{bmatrix}
        x_1z_1 & x_1z_2 & ...    & x_1z_M \\
        x_2z_1 & x_2z_2 & ...    & x_2z_M \\
        \vdots   &          & \ddots & \vdots \\
        x_Nz_1 & x_Nz_2 & ...    & x_Nz_M \\
    \end{bmatrix}
\label{eq:concat}
\end{equation}

Results of these fusion models can be seen in Table \ref{tab:fusion_results} and Fig.~\ref{fig:prec_recall_deep}. 
The results demonstrate that the proposed method achieves the highest recall and precision rate among alternative fusion methods. Additionally, we observe that Orderless Bilinear model performs worse than the Concatenation. We believe that summation approach over all the spatio-temporal indices in the former model loses the information instead of achieving fusion.

\begin{table}
\begin{footnotesize}
\begin{center}
  \begin{tabular}{| l || c | c | c | }
    \hline
    Model        & Recall & Precision & F1-Score\\ \hline \hline
    Average Fusion \cite{simonyan2014two,karpathy2014large}          & 0.90 & 0.92 & 0.91 \\ \hline
    Dempster Shafer Fusion \cite{lee2017fusion}         & 0.93 & 0.95 & 0.94 \\ \hline
    Concatenation-FC \cite{socher2012convolutional,karpathy2014large}          & 0.91 & 0.89 & 0.90 \\ \hline
    OP-Concatenation  & 0.93 & 0.90 & 0.91 \\ \hline
    Orderless Bilinear \cite{lin2015bilinear}      & 0.87 & 0.90 & 0.88 \\ \hline
    \textbf{OP-Bilinear Fusion}  & \textbf{0.97} & \textbf{0.96} & \textbf{0.96} \\ \hline
    
  \end{tabular}
\end{center}
\caption{Precision, Recall and F1-Score values for different fusion methods and proposed method. Cited papers use similar (multi-modal or feature) fusion methods to our experimentation models.}
\label{tab:fusion_results}
\end{footnotesize}
\end{table}


\subsection{Impact of 3D Convolutions}

In this section we investigate the merits of 3D convolutions. Since the model is order preserving (OP), output of the fusion model is a spatio-temporal tensor. This tensor allows us to leverage 3D convolutions to reduce the total number of parameters and chances of overfitting by exploiting spatio-temporal correlations. We demonstrate this by comparing OP models that exploit 3D convolutions with corresponding fully connected models on two different fusion approaches, i.e., concatenation and bilinear feature descriptors. Table \ref{tab:order_results} demonstrates that models that preserve order achieve superior performance in both fusion approaches.

\begin{table}
\begin{footnotesize}
\begin{center}
  \begin{tabular}{| l || c | c | c | }
    \hline
    Model       & Recall & Precision & F1-Score\\ \hline \hline
    Concatenation-FC           & 0.91 & 0.89 & 0.90 \\ \hline
    OP-Concatenation & 0.93 & 0.90 & 0.91 \\ \hline\hline
    Bilinear-FC & 0.95 & 0.75 & 0.85 \\ \hline
    \textbf{OP-Bilinear Fusion} & \textbf{0.97} & \textbf{0.96} & \textbf{0.96} \\ \hline
    
  \end{tabular}
\end{center}
\caption{Precision, Recall and F1-Score values for Order Preserving (OP) fusion methods and their fully connected orderless variants. OP methods exploit 3D convolutions, other methods do not.}
\label{tab:order_results}
\end{footnotesize}
\end{table}


\textbf{Order Preserving Concatenation: }In this model, we adjust our order preserving approach to concatenation methods. We concatenate the features from each modality at every spatio-temporal index as in Eq.~\ref{eq:concat_vector}. However, instead of stacking the vectors further (as in Eq.~\ref{eq:concat_os}), we use these concatenated vectors as spatio-temporal local descriptors with $M+N$ length  feature vector $\bm{o}_{u, v, t}$ at each index ($u,v,t$). Since the spatio-temporal order of descriptors is preserved this allows us to use 3D convolutions to exploit correlations. Tables \ref{tab:fusion_results}, \ref{tab:order_results} show the results of this model under the label `OP-Concatenation' and demonstrates that order preserving concatenation performs better than simple concatenation.


\textbf{Bilinear-Fully Connected: }In this model, we replace the 3D convolutions from the model in Section \ref{sec:bilinear} with fully connected layers and fine-tune the network similarly with pre-trained CNN weights. This effectively removes the weight sharing of 3D convolutions, which removes the order-preserving aspect of the model and makes the model prone to overfitting. 

Table \ref{tab:order_results} demonstrates the improvement in performance with preserving order on Bilinear Feature descriptors. OP-Bilinear model results with significantly fewer false positive rates, i.e, much higher precision compared to fully-connected method.

\subsection{Averaging OF and LSTM Comparison}\label{sec:lstm}

Our visual input is the magnitudes of OF vectors averaged over a time interval. Extracting OF from low-resolution cameras generate noisy inputs. Additionally, for this application, location and existence of the motion is as important as the evolution of the motion. Spatial location of the motion captured among subsequent frames does not change drastically and averaging over a short time interval allows OF magnitudes to compress the motion captured while reducing the noise. This generates a low dimensional, compact feature description. However, a more complex and higher dimensional approach is capable of an incrementally better performance. Recurrent Neural Networks (RNNs) and Long-Short Term Memory (LSTMs) models have been shown to achieve good performance on variety of tasks \cite{lrcn2014,srivastava2015unsupervised,yue2015beyond}. We compare the performance of our averaged OF model with an LSTM model. In the LSTM model each input frame (OF Magnitude) goes through the convolutional part of the 'Visual' model from Section \ref{sec:single_CNN} and the outputs of the consecutive frames are fed into an LSTM cell similar to Activity Recognition model in \cite{lrcn2014}. Table \ref{tab:lstm_results} shows the performance of the LSTM compared to averaged OF visual model. This demonstrates that averaging reduces the dimensionality and has slightly better false positive rates compared to small improvement in detection performance of LSTMs. Additionally, for low-power strategic scenarios, processing every frame through a CNN model may not be possible(which is required in LSTM) whereas taking an average over a time interval and processing only this compact snapshot is more feasible.

\begin{table}
\begin{small}
\begin{center}
  \begin{tabular}{| l || c | c | c | }
    \hline
    Model          & Recall & Precision & F1-Score\\ \hline \hline
    Visual    & 0.82 & 0.89 & 0.86 \\ \hline
    LSTM        & 0.86 & 0.86 & 0.86 \\ \hline

  \end{tabular}
\end{center}
\caption{Comparison of the visual and LSTM model.}
\label{tab:lstm_results}
\end{small}
\end{table}

\section{Conclusions}
In this work, we introduced an OP-Bilinear Fusion method to jointly leverage sensor data and imagery. By conducting a series of experiments we analyzed the impact of each module.  We demonstrated that our feature selection algorithm makes the fusion method feasible by effectively reducing dimensionality with only a small tradeoff in single modality detection performance. We showed that our fusion model performs improves performance over models trained on single modalities and demonstrated that the fusion is beneficial. We compared the proposed fusion method with the traditional multi-modal and feature fusion methods and achieved better performance with the proposed method. Finally, we compared our approach of averaging OF frames to a more complicated LSTM approach and showed that by averaging multiple OF frames the sequence information is not lost and the model performs similarly.


The proposed method demonstrates the benefits of retaining the order when using a bilinear operator with video and seismic signals. However, the principle of preserving structural order with bilinear operators may be extended to any combinations of spatial or temporal data sources since the formulation in Eq. \ref{eq:output_long} is generic.
Furthermore, by replacing the outer product operation with tensor product operation, method can be expanded for more than two modalities. 
where tensor product of three feature vectors can be expressed as $\bm{T} = \bm{a} \otimes \bm{b} \otimes \bm{c}$ where $T_{i,j,k} = a_ib_jc_k$. 


\section{Acknowledgements}
The authors thank Dr. Thyagaraju Damarla at Army Research Lab for providing the dataset and guidance in processing the data.
Research was supported in part by the Army Research Laboratory and was accomplished under Cooperative Agreement Number W911NF-09-2-0053 (the ARL Network Science CTA). The views and conclusions contained in this document are those of the authors and should not be interpreted as representing the official policies, either expressed or implied, of the Army Research Laboratory or the U.S. Government. The U.S. Government is authorized to reproduce and distribute reprints for Government purposes notwithstanding any copyright notation here in.

{\small
\bibliographystyle{ieee}
\bibliography{fusion_ulutan}
}

\end{document}